\documentclass[a4paper, 10pt, conference]{ieeeconf}      % Use this line for a4 paper

\IEEEoverridecommandlockouts                              % This command is only needed if 
\usepackage{amsmath} % assumes amsmath package installed
\usepackage{multicol} % assumes amsmath package installed
\usepackage{multirow} % assumes amsmath package installed
\usepackage{relsize}
\usepackage[caption=false,captionskip=-5pt]{subfig}
\usepackage{cite}
\usepackage{graphicx}
\usepackage{hhline}
\usepackage{algorithm}
\usepackage[noend]{algpseudocode}
\usepackage{float}
\usepackage[table]{xcolor}
\usepackage{url}
\usepackage{hyperref}

\algnewcommand{\IfThenElse}[3]{% \IfThenElse{<if>}{<then>}{<else>}
  \State \algorithmicif\ #1\ \algorithmicthen\ #2\ \algorithmicelse\ #3}

\title{\LARGE \bf
Dual-Arm Construction Robot for Automatic Fixation of Structural Parts to Concrete Surfaces in Narrow Environments
}

\author{Andr\'e Yuji Yasutomi$^*$, Toshiaki Hatano$^*$, Kanta Hamasaki, Makoto Hattori, and Daisuke Matsuka% <-this % stops a space
\thanks{Andr\'e Yuji Yasutomi, Toshiaki Hatano, Kanta Hamasaki, Makoto Hattori, and Daisuke Matsuka are with the Robotics Research Department, Controls and Robotics Innovation Center, R\&D Group, Hitachi, Ltd. {$^{1}$\tt\footnotesize andre.yasutomi.ss@hitachi.com}}%
\thanks{Digital Object Identifier (DOI): \href{https://doi.org/10.1109/SII55687.2023.10039387}{10.1109/SII55687.2023.10039387}}
}

\pubid{
\begin{tabular}[t]{@{}c@{}}~\copyright~2023 IEEE. Personal use of this material is permitted. Permission from IEEE must be obtained for all other uses, \\
                                    in any current or future media, including reprinting/republishing this material for advertising or promotional purposes, creating new collective works, \\ for resale or redistribution to servers or lists, or reuse of any copyrighted component of this work in other work\end{tabular}}

\begin{document}
\maketitle

\begingroup\renewcommand\thefootnote{*}
\footnotetext{These authors contributed equally.}
\endgroup

%%%%%%%%%%%%%%%%%%%%%%%%%%%%%%%%%%%%%%%%%%%%%%%%%%%%%%%%%%%%%%%%%%%%%%%%%%%%%%%%
\begin{abstract}
Fixation of structural parts to concrete is a repetitive, heavy-duty, and time-consuming task that requires automation due to the lack of skilled construction workers. Previously developed automation techniques have not achieved the complete fixation of structural parts  and are  difficult to implement in narrow construction environments. In this study, we propose  a construction robot system that enables the complete installation of structural parts to concrete and can be easily introduced to unstructured and narrow construction environments. The system includes two arms that simultaneously position and fix the structural parts, and custom tools that reduce the reaction force applied to the robots so that smaller robots can be used with lower payloads. Due to the modular design of the proposed system, it can be transported in parts for easy introduction to the construction environment. We also propose a procedure for fixing structural parts. Experimental results demonstrate that the custom tools make it possible to use smaller robots without moment overload in the robot joints. Moreover, the results show that the proposed robot system and fixation procedure enable automatic fixation of a structural part to concrete.
\end{abstract}
 
%%%%%%%%%%%%%%%%%%%%%%%%%%%%%%%%%%%%%%%%%%%%%%%%%%%%%%%%%%%%%%%%%%%%%%%%%%%%%%%%
\section{INTRODUCTION}

On-site construction involves dirty, dangerous, and difficult manual tasks performed outdoors \cite{construction1}. This line of work has been losing popularity among young workers, which is causing a labor shortage in the construction industry \cite{construction2, construction_shortage}. Additionally, due to the lack of young laborers, the skills required for construction tasks are not being passed  onto the next generations, which may lead to a collapse in the industry \cite{construction1,construction_us}. To avoid this, efforts have been made to automate construction tasks using industrial robots, as they are suitable for tasks that are dull, repetitive, and require high precision.

One task that is frequently conducted  in construction is the installation of structural parts such as mechanical, electrical, and plumbing (MEP) support systems in concrete surfaces. This task involves positioning the structural parts, drilling holes, inserting anchor bolts into the holes, and tightening nuts to fix the structural parts. Examples of such structural parts are brackets and hangers which are normally fixed to concrete walls and ceilings.

\begin{figure}[tb]
  \begin{center}
  \includegraphics[width=\columnwidth]{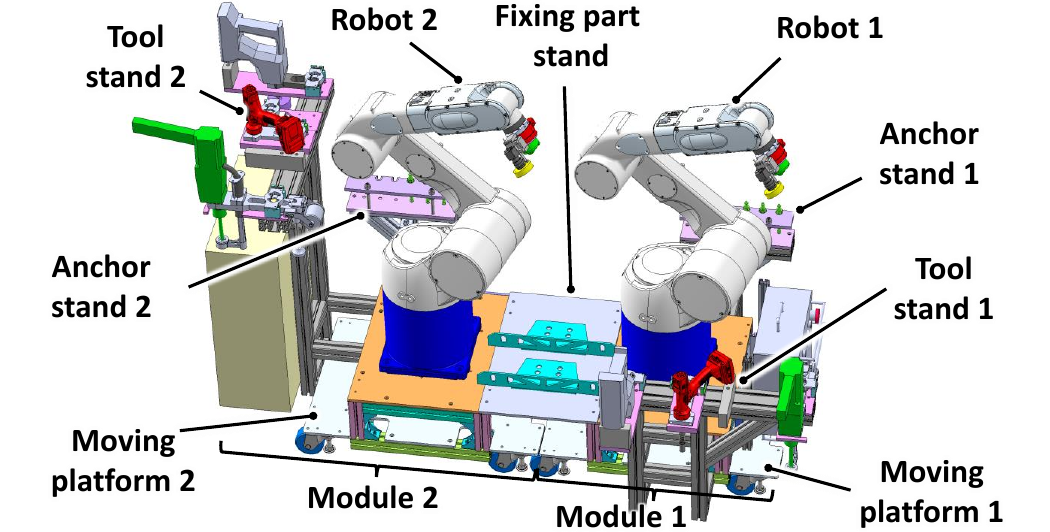}
  \caption{Conceptual design of structural part fixing system for concrete surfaces.}
  \label{fig:entire_system}
  \end{center}
\end{figure}
 
To automate this task, robots that perform hole drilling \cite{jaibot,schindler_tall_bldgs} and anchor bolt insertion into concrete surfaces \cite{schindler_tall_bldgs,andre_icra2021} have been developed . Xu et al. \cite{jaibot} introduced a robot that automatically drills holes into concrete ceilings for fixing MEP support systems. The robot consists of a single robot arm on top of a crawler robot that drills holes on the basis of building information modeling (BIM)  data. Balzan et al. \cite{schindler_tall_bldgs} presented an elevator installation system consisting of a robot on top of a lifted platform that drills holes in the concrete walls of elevator shafts and inserts anchor bolts into the holes. Brackets are then fixed by humans for rail installation. In a prior study  \cite{andre_icra2021}, we proposed a deep reinforcement learning-based anchor bolt insertion strategy for insertion even when positioning is imprecise due to oscillation in the robot base and tool looseness. Though these solutions have enabled hole drilling and anchor insertion, they do not complete the structural part fixation. They also use a single robot for the hole procedure, which makes it difficult to position and fix the structural part at the same time. Moreover, in the case of \cite{jaibot} and \cite{schindler_tall_bldgs}, the robots were large (length$>$1400 mm) and heavy ($>$100 kgs). This makes it difficult for them to work in narrow construction environments (minimum of 1.5 $\times$ 2.5 m) and to be transported to the work environment, as the delivery routes consist of narrow corridors (width$<$2 m).

\pubidadjcol

The use of two arms to automate assembly tasks in construction is not new; a  robotic systems was previously developed for installing ceiling boards and raised floors in construction sites \cite{shimizu_robots}. This robotic system consists of one robot that positions objects to be fixed at locations determined by measurements taken from image and laser sensors, and another robot that fixes the objects by screwing bolts. This system enables complete fixation of structural parts; however, the robots do not fix these parts into concrete or conduct heavy-duty tasks such as hole drilling. Moreover, the system also uses large and heavy robots which are placed on a long moving platform (2.5 m). This platform is difficult to transport through the narrow corridors of construction sites as it can get stuck in the corners.

In this study, we propose a robotic system (Fig. \ref{fig:entire_system}) that enables the complete installation of structural parts to concrete and can be easily introduced to unstructured and narrow construction environments.. The system includes two arms that position and fix the structural parts at the same time, and custom tools that reduce the reaction force applied to the robots so that smaller robot can be used with lower payloads. Due to its modular design, the proposed system can be transported in parts for easy introduction to the construction environment. 

The main contributions of this study are as follows: (1) A robotic system design  for complete installation of structural parts and easy introduction to the construction field; (2) Custom tools that reduce reaction force to enable smaller robots (low payload) to perform heavy-duty tasks; (3) A procedure for efficiently fixing structural parts in unstructured environments.

\section{PROPOSED DESIGNS AND PROCEDURE}\label{sec:methods}
\subsection{Structural part fixing system design}

Fig. \ref{fig:entire_system} presents the design proposed to automatically fix structural parts to concrete. The proposed system is composed by two industrial robots, each of which is provided with its own tool and anchor stand. This independent modular structure enables the robots to execute tasks without sharing the same workspace, enabling them to work in parallel and, thus, increase throughput. This structure also enables the robots to be separated for easy introduction to the construction field without the need for adjusting the program for tool change before the fixation task is executed. The only adjustment required due to the separation is when the robots interact with each other or with the environment, which is addressed by image and laser measurements of target objects with the camera and laser sensor each robot is equipped.

The system also includes a fixing part stand where the parts to be fixed are stored and a moving platform per module on which all of the robot components are installed. Each tool stand contains a hole drilling tool, an anchor hammering tool, and a nut tightening tool. Tool stand 2 also includes a gripper which is used to pick and place the part to be fixed. A single gripper is used because there is no need for both robots to pick up the part at the same time. If a part to be fixed has more than three fixation points, both robots can work in parallel to drill holes, insert anchors, and tighten the nuts after one point has already been fixed (as will be detailed in section \ref{subsec:procedure}). 

It is important to note that the system design proposed in this study is only feasible if arms shorter than 1.3 m are used because it would be difficult to operate two arms longer than that in narrow construction environments. However, because arms shorter than 1.3 m normally only support low payloads (under 15 kg), custom tools are needed to minimize the reaction force applied to the robot in order to carry out heavy-duty tasks. This motivated the development of the custom tools which will be described in the next section.

\begin{figure}[t]
  \begin{center}
  \includegraphics[width=\columnwidth]{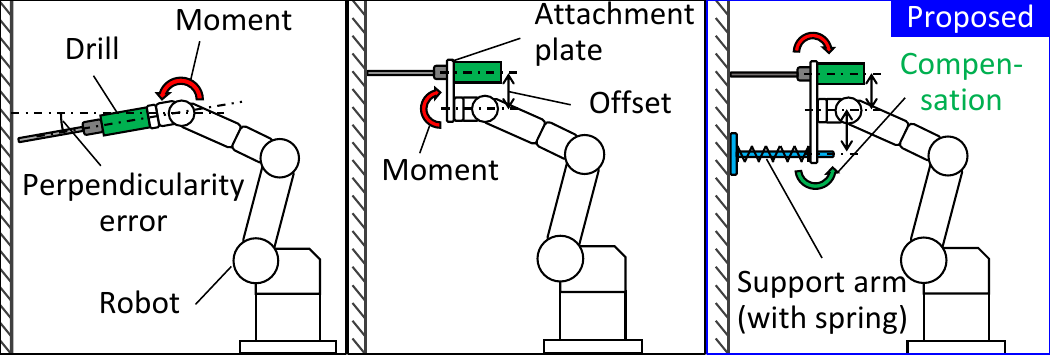}
  \captionsetup{justification=centering}
  \subfloat[\label{subfig:drillconcept1}Aligned axis\\(Conventional)]{\hspace{.35\columnwidth}}
  \subfloat[\label{subfig:drillconcept2}Offset axis\\(Tool shortened)]{\hspace{.3\columnwidth}}
  \subfloat[\label{subfig:drillconcept3}Moment compensated\\(Proposed)]{\hspace{.35\columnwidth}}
  \caption{Hole drilling tool designs considered.}
  \label{fig:drill_design}
  \end{center}
\end{figure}

\begin{figure}[t]
  \begin{center}
  \includegraphics[width=\columnwidth]{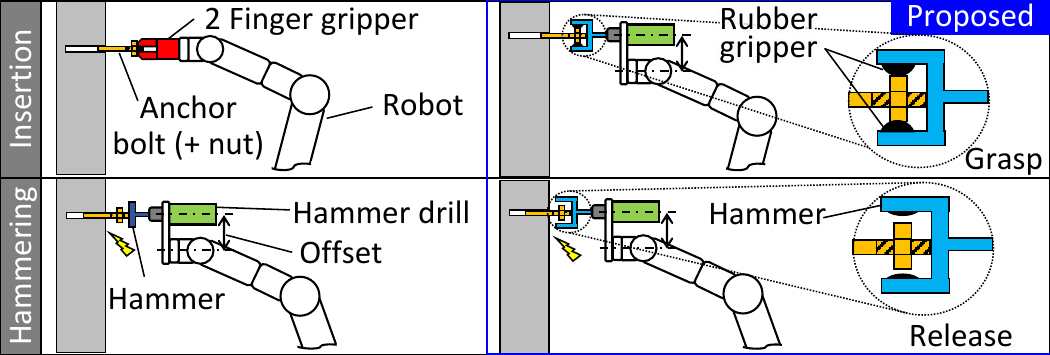}
  \subfloat[\label{subfig:hamconcept1}Grasp/hammering separate]{\hspace{.45\columnwidth}}
  \subfloat[\label{subfig:hamconcept2}Grasp/hammering together]{\hspace{.6\columnwidth}}
  \caption{Hole drilling tool designs considered.}
  \label{fig:ham_desing}
  \end{center}
\end{figure}

\subsection{Custom tool designs}\label{subsec:holeseach}
\subsubsection{Hole drilling tool} \label{subsubsec:drill_design}
To drill holes into concrete surfaces in narrow spaces while reducing the reaction forces applied to the robot, we considered the hole drilling tool designs shown in Fig. \ref{fig:drill_design}. The aligned axis approach (Fig. \ref{subfig:drillconcept1}) is the simplest approach and has been used in prior studies \cite{jaibot,schindler_tall_bldgs}. However, this approach results in a long end effector which is difficult to control in narrow spaces. Moreover, due to the long distance between the drill tip and the tool fixing point, small perpendicularity errors between the tool and the target surface generate a high moment in the robot joints, which causes moment overload. The offset axis approach shown in Fig. \ref{subfig:drillconcept2} is a possible solution to this problem. This approach involves fixing the tool to  an attachment plate that offsets the drill to place it parallel to the robot flange. Although the end effector can be shortened, it still generates an undesired moment in the robot joints due to the axis offset. To address this, we propose the design shown in Fig. \ref{subfig:drillconcept3}. The proposed design is similar to the offset axis approach, but it contains a support arm that compensates for the moment generated by drilling. The support arm includes a linear guide that enables a rod (with increased wall contact surface) to move perpendicular to the target surface. The rod is actuated by a spring that passively applies the compensation force on the wall and also enables the rod to return to its initial length after hole drilling. The spring is a constant load spring because a regular spring increases the compensation moment as the drill reaches longer depths. This causes overcompensation of the drill moment, which is an issue that will be detailed in section \ref{subsec:drill_res}

\begin{figure*}[tb]
  \begin{center}
  \includegraphics[width=\linewidth]{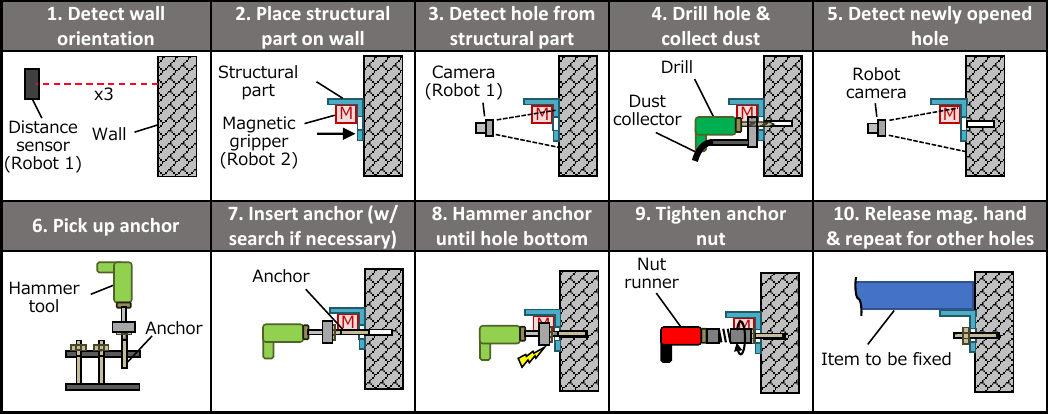}
  \caption{Procedure for fixing structural parts to concrete walls.}
  \label{fig:procedure}
  \end{center}
\end{figure*}

\subsubsection{Anchor hammering tool}
For the anchor hammering tool, the designs shown in Fig. \ref{fig:ham_desing} were considered. Both designs follow the offset approach of the proposed drilling tool to reduce the tool length. However, the moment was not compensated for because preliminary results showed that anchor hammering requires less force. The reaction force for this tool was only compensated for by adding an elastic material between the attachment plate and the robot flange, which reduces the high frequency oscillations generated during anchor hammering. 

The main design consideration for this tool was the grasping method. While the approach shown in Fig. \ref{subfig:hamconcept1}, which uses separate tools for anchor bolt grasping (for initial insertion) and hammering, is the simplest in terms of tool complexity, it requires the development of two different tools, space in the tool stand to store the two tools, and more time to execute the task because the tool needs to be changed. To avoid this, we devised the tool shown in Fig. \ref{subfig:hamconcept2}. This approach combines the anchor grasping and hammering into a single tool. The proposed tool includes a cup-shaped hammer that is equipped with an inflatable rubber gripper. This gripper is inflated to grasp the anchor bolt from the nut for insertion and is deflated after insertion for hammering with the hammer's flat surface inside the cup. Note that by grasping the anchor with the nut included, the task of screwing the nut into the anchor bolt can be also avoided, which further shortens the task execution time.

\subsubsection{Nut tightening tool}\label{subsubsec:nutrun}
Following the designs of the drilling tool and the anchor hammering tool, the nut tightening tool design involves attaching a nut runner offset from the robot flange so that the tool is shorter and easy to manipulate in narrow spaces. By fixing the tool offset from the robot flange, the reaction moment around the flange axis can be reduced as the moment caused by the nut runner tightening is offset from the flange axis. As a result, the reaction moment is divided into the moment at the flange axis, and of the force applied to the attachment plate which is used as lever.

% \subsubsection{Gripper}

\subsection{Fixation procedure}\label{subsec:procedure}
We devised the procedure shown in Fig. \ref{fig:procedure} for carrying out the complete fixation of the structural part to concrete. The procedure consists of the following steps: (1) robot 1 detects the target surface orientation with a distance sensor in order to estimate the robot system orientation relative to the target surface in order execute the task perpendicular to this surface; (2) robot 2 picks up the structural part and places it on the target surface (e.g., wall or ceiling) at a position pre-determined by, for example, BIM data; (3) robot 1 uses its  camera to detect the hole of the structural part where the anchor bolt should be inserted; (4) robot 1 attaches the drilling tool to its end, approaches the detected structural part hole position, and drills a hole in the wall through the structural part hole while collecting dust with the dust collector; (5) robot 1 detaches the tool and detects the newly drilled hole in the wall with the camera; (6) robot 1 attaches the anchor hammering tool and picks up the anchor by inflating the rubber gripper; (7) robot 1 approaches the drilled hole position detected by the camera and attempts anchor bolt insertion. If the anchor is not inserted in the first try, a hole search is conducted with either a model-based \cite{pih_modelbasedsearch} or a deep learning-based strategy \cite{andre_icra2021,andre_sii2022}; (8) after the anchor is partially inserted (until its maximum allowed depth without hammering), robot 1 deflates the rubber gripper to release the anchor, and the hammer tool is activated to hammer the anchor until it reaches the bottom of the hole; (9) robot 1 changes tools to the nut runner and approaches the anchor position which is the same as the hole position previously detected by the camera. Then the nut runner tightens the nut of the anchor bolt to fix one side of the structural part; (10) robot 2 releases the structural part, and steps 3 to 9 are repeated for the other holes of the structural part with either of the robots.

In this procedure, step 1 is carried out  because the robot base may not necessarily be perpendicular to the walls (or parallel to the ceiling, in the case of drilling in the ceiling) due to the uneven ground at construction sites. Although the structural part hole can be estimated from CAD data and the end-effector position of robot 2, slippage during robot grasping or misalignment between robots could generate a position error that causes the drill to touch the structural part surface, damaging both the structural part and the drill bit. The purpose of step 3 is to avoid this problem. Drilling holes through the part holes in Step 4 avoids the need to align the hole drilled in the wall with the holes of the structural parts, in which small errors may make it impossible to carry out step 7. Other options could be to drill the holes and hammer the anchors before placing the structural part, which is the procedure conducted by humans. However, in this case, the anchor nut would have to be picked and fit to the anchor bolts afterwards, which is a challenging task and requires the nuts to be placed separately from the anchors in the anchor stand. The proposed procedure picks the anchors with the nuts specifically to avoid this task. Lastly, the purpose of step 5 is to prevent errors generated between the target and the actual hole drilling positions resulting from the looseness of the drill bit of the drilling tool and the flexibility of the attachment plate. This proposed procedure enables a robot to efficiently fix structural parts to concrete surfaces.

\section{IMPLEMENTATION}\label{sec:setupandconditions}
\subsection{Experimental setup}\label{subsec:setup}
\begin{figure}[tb]
  \begin{center}
  \includegraphics[width=\columnwidth]{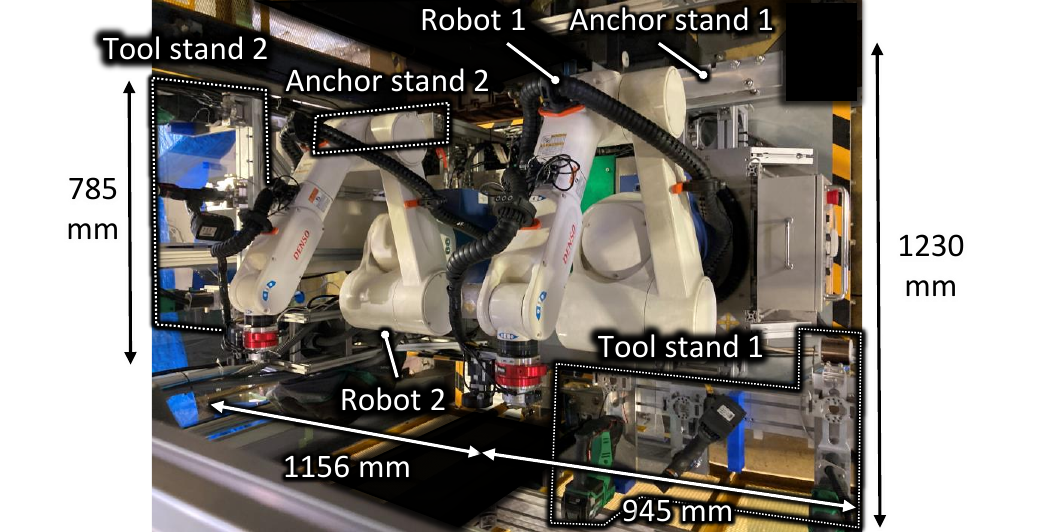}
  \caption{Experimental setup (top right view).} 
  \label{fig:setup}
  \end{center}
\end{figure}

To test the proposed custom tools and the structural part fixing procedure, the experimental setup shown in Fig. \ref{fig:setup} was implemented. The setup includes two Denso robots VM-60B1 which are 1110 mm in length (with a reach of 1298 mm), and 82 kg in weight, with a 13-kg payload. These robots are small compared to the arms used in previous approaches \cite{jaibot,schindler_tall_bldgs,shimizu_robots} (length$>$1.2 m, payload$>$14 kg, weight$>$200 kg/robot). The setup shares the same structure as the system shown in Fig. \ref{fig:entire_system}. The entire system size has a maximum length (L), width (W), and height (H) of 2105, 1265, 1950 mm, respectively, but during transportation, module 1 is L945xW1230xH1220 mm and module 2 is L1156xW785xH1220 mm. This size can be easily transported through narrow corridors and is considerably shorter in length than \cite{schindler_tall_bldgs,shimizu_robots} (L$>$2500 mm) and in height than \cite{jaibot,schindler_tall_bldgs,shimizu_robots} (min H$>$1800 mm). The setup also includes a concrete block (size: H300xW200xT150 mm, compressive strength: 24 N/mm\textsuperscript{2}) placed close to robot 1 so the robot can test the proposed procedure by fixing one side of a structural part to the block. The setup also includes the tools for part fixing, the details of which are shown in section \ref{subsec:tools}.

Each robot is equipped with a force and torque (FT) sensor (WACOH-TECH Dyn Pick WEF-6A1000-30-RCD-B), a laser distance sensor (Keyence IL-300), a tool changer (Denso AHC5), and a camera (Sharp IV-S300CE) with a controller (IV-S300J) that includes an image processing engine. The FT sensor is used to measure the force and torque at the robot flange. This measurement makes it possible to detect when the tool touches the target surface and when the robot stops in the event of force or moment overload (maximum of \textpm1000 N and \textpm30 Nm, respectively). The laser distance sensor enables the measurement of the robot distance to the concrete block. This is used to monitor tool or anchor depth inside the wall during drilling or anchor insertion, respectively, and to estimate the wall orientation with the method to be described in section \ref{subsubsec:wall_orient}. We observed that monitoring the tool and anchor depth with the laser sensor is more accurate than simply monitoring the robot end-effector position because the slippage of the moving platform toward the opposite direction from the target surface during task execution generates differences between the end-effector position, the actual tool, and the anchor depth reached. The tool changer enables the tool to be attached and detached through chucks opened and closed by compressed air. The image processing engine of the camera controller makes it easy to configure traditional algorithms for detection of targets which is detailed in section \ref{subsubsec:detection}.

Custom programs were written in the Denso Robot language (PAC Script) to control the robot. The programs were produced in a way that the robot trajectories could adapt to the detected wall orientations and target positions such as structural part hole, wall hole, and anchor bolt positions. The programs were written in a modular structure so the execution order could be easily changed. 

\subsection{Tools}\label{subsec:tools}
\subsubsection{Hole drilling tools} 
\begin{figure}[tb]
  \begin{center}
  \includegraphics[width=\columnwidth]{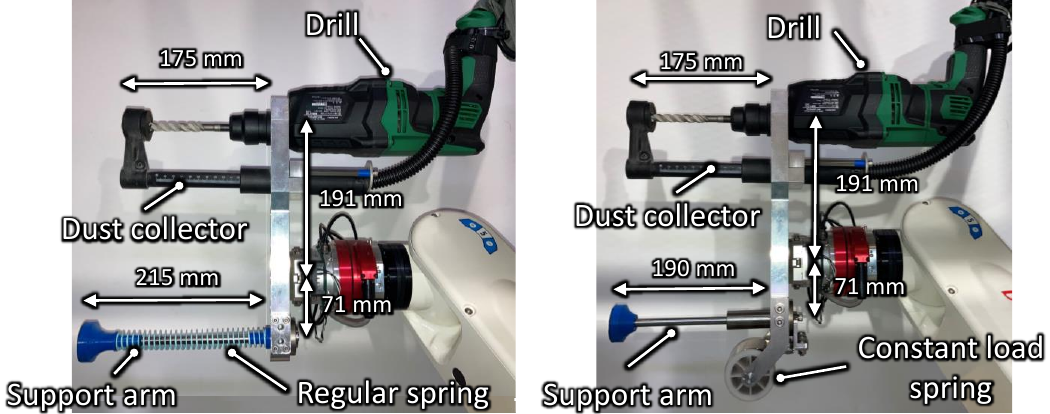}
  \subfloat[\label{subfig:drill_imp1}Tool 1: with regular spring]{\hspace{.5\columnwidth}}
  \subfloat[\label{subfig:drill_imp2}Tool 2: with constant load spring]{\hspace{.5\columnwidth}}
  \caption{Implemented hole drilling tools.}
  \label{fig:drills}
  \end{center}
\end{figure}

The hole drilling tools implemented are shown in Fig. \ref{fig:drills}. Drilling tool 1 was implemented with a regular spring (Misumi SWU21-175, k=2150 N/m) to prove that it causes the moment overcompensation issue mentioned in section \ref{subsubsec:drill_design}. Drilling hole 2 was implemented with a constant load spring (Samini TCL15, F=147 N) to realize the proposed design and show that it provides a fixed compensation moment which does not cause moment overcompensation. Both tools are equipped with a Hikoki DH28PCY2 drill, and drill bits 12 mm in diameter and 160 mm in length.

\subsubsection{Anchor hammering tool} 
\begin{figure}[tb]
  \begin{center}
  \includegraphics[width=\columnwidth]{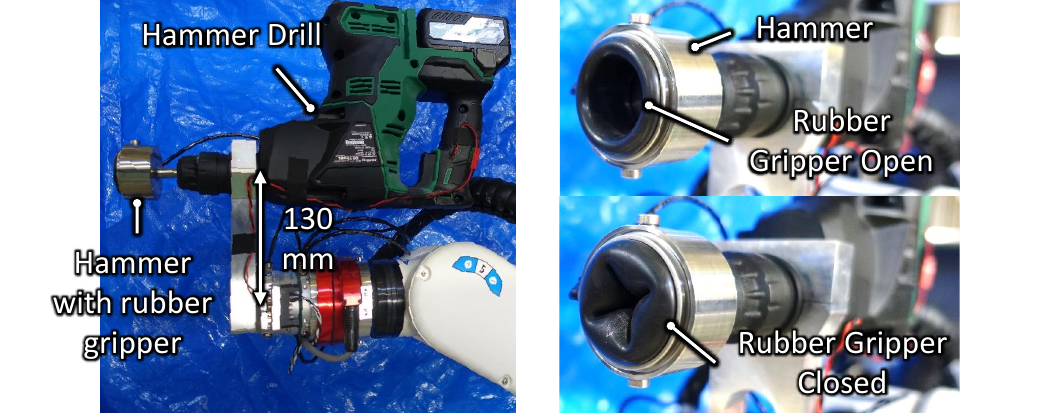}
  \caption{Implementation of hammer tool.}
  \label{fig:hammer}
  \end{center}
\end{figure}

The hammering tool implemented is shown in Fig. \ref{fig:hammer}. This tool was equipped with a rubber gripper (Bridgestone U030GCF) that, when inflated at a pressure of 0.15 MPa, enables the tool to grasp the anchor bolt from the nut (type: wedge, diameter: 12 mm, length: 126 mm, mass: 113 g). The rubber gripper was fixed to the hammer tool through a hammer which was used to hammer the tip of the anchor bolt. This hammer was made of pre-hardened steel to avoid deformation during anchor hammering. The hammer was driven by a hammer drill (Hikoki DH 18DBL) which was set to perform a hammering movement without rotation.

\subsubsection{Nut tightening tool} Fig. \ref{fig:nutmag} shows the nut tightening tool implemented in this study. A nut runner (Desoutter BLRTC045-3990-10S) was fixed to the tool offset from the robot flange. This nut runner is provided with a pulse mechanism which, like the design proposed in section \ref{subsubsec:nutrun}, also reduces the reaction moment transmitted to the robot during tightening. A custom trigger is installed in the nut runner to enable the robot to activate the tool. The nut runner  is also provided with a socket that contains a spring that enables the tightening of the anchor bolt nut to a certain extent without the robot having to actively advance toward the anchor axis to account for the nut advance as it is tightened. This design simplifies the nut tightening control.

\subsubsection{Gripper} 
\begin{figure}[tb]
  \begin{center}
  \includegraphics[width=\columnwidth]{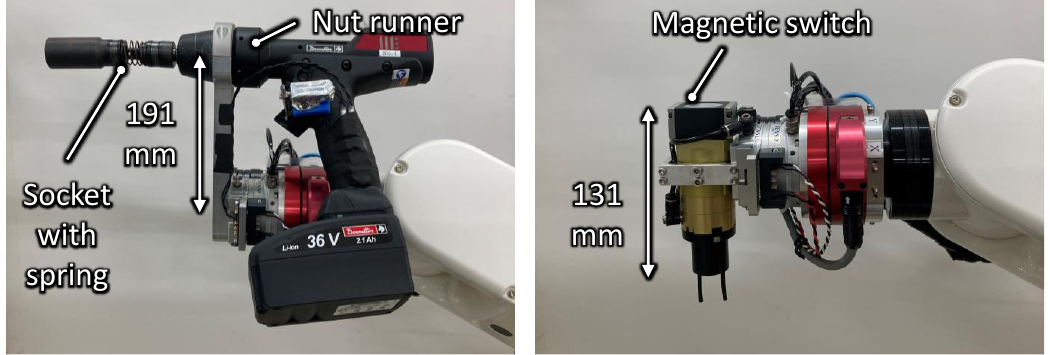}
  \caption{Nut tightening tool and structural part gripper implementations.}
  \label{fig:nutmag}
  \end{center}
\end{figure}

The gripper used in this study is also shown in Fig. \ref{fig:nutmag}. It consists a magnetic switch (Magswitch AR30) attached to a developed jig that connects the switch to the tool changer. The magnetic switch is turned on and off by compressed air and makes it possible to grasp the structural part which is made of steel.

\subsection{Adaptation to unstructured environment}
\subsubsection{Wall orientation estimation}\label{subsubsec:wall_orient}
The wall orientation was estimated by the following steps: (1) moving the robot to close to the wall and measuring the distance with the laser sensor, (2) moving the robot a pre-determined distance to the right in the robot xy plane and measuring the distance again, (3) moving the robot a pre-determined distance toward the bottom of the first measurement point in the same plane and measuring the distance again, and (4) using the following equations to determine the wall coordinate system with the method presented in a prior study \cite{orient_calc}.
\begin{align}
  \hat{x}&=\frac{P_2-P_1}{||P_2-P_1||} \\ 
  \hat{z}&=\frac{\hat{x} \times (P_3-P_1)}{||\hat{x} \times (P_3-P_1)||} \\
  \hat{y}&=\hat{z}\times \hat{x}
\end{align}

Here, $P$ are the positions of the measured points, and $\hat{x}$, $\hat{y}$, $\hat{z}$ are the versors of the wall coordinate system. The robot movements and detections followed this coordinate system.

\subsubsection{Hole detection}\label{subsubsec:detection}
\begin{figure}[tb]
  \begin{center}
  \includegraphics[width=\columnwidth]{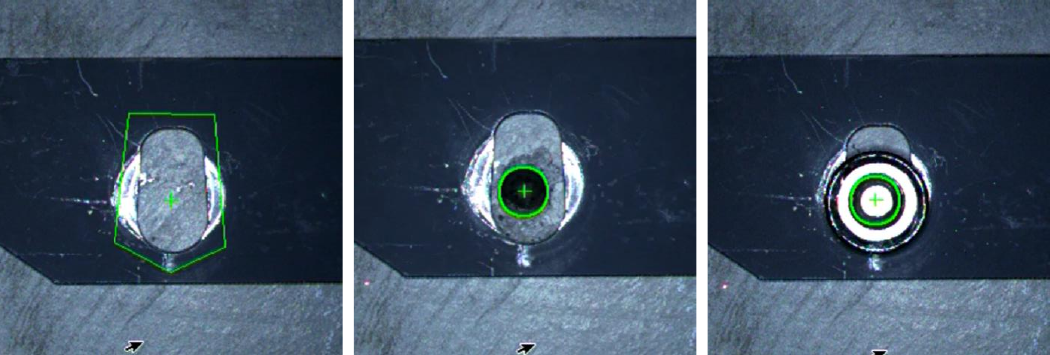}
  \subfloat[\label{subfig:detect1}Part hole detection]{\hspace{.33\columnwidth}}
  \subfloat[\label{subfig:detect2}Wall hole detection]{\hspace{.33\columnwidth}}
  \subfloat[\label{subfig:detect3}Anchor bolt detection]{\hspace{.33\columnwidth}}
  \caption{Detection of support hole and newly drilled hole in concrete.} 
  \label{fig:detect}
  \end{center}
\end{figure}

Hole detection was performed by the image processing engine contained in the controller of the camera used (see section \ref{subsec:setup}). The engine includes multiple ``modules'' that performed detections using model-based algorithms. The ``multiple search module'' (Fig. \ref{subfig:detect1}) is used to search for the structural part hole, in which algorithm compares the current image with multiple previous captured images of the structural part similar to template matching algorithms \cite{template}. The ``shape module'' is used for wall hole (Fig. \ref{subfig:detect2}) and anchor bolt (Fig. \ref{subfig:detect3}) detection, which searches for circles that are similar in size to the circle in a template image (similar to shape matching algorithms \cite{shape}).

\section{EXPERIMENTAL RESULTS}\label{sec:results}
\subsection{Drilling measurement} \label{subsec:drill_res}
To evaluate the drilling tools implemented, hole drilling was attempted with both tools implemented, as well as with a tool without the support arm (no compensation). The tests were conducted in the following steps: (1) moving the tool with simple position control toward the wall until the FT sensor detected that the tool touched the wall, (2) activating the drill, and (3) continuing the movement at a speed of 2.25 mm/s until the tool reached a depth of 80 mm inside the wall (detected by the distance sensor) or until the moment limit is reached (\textpm30 Nm). 

\begin{figure}[tb]
  \begin{center}
  \includegraphics[width=\columnwidth]{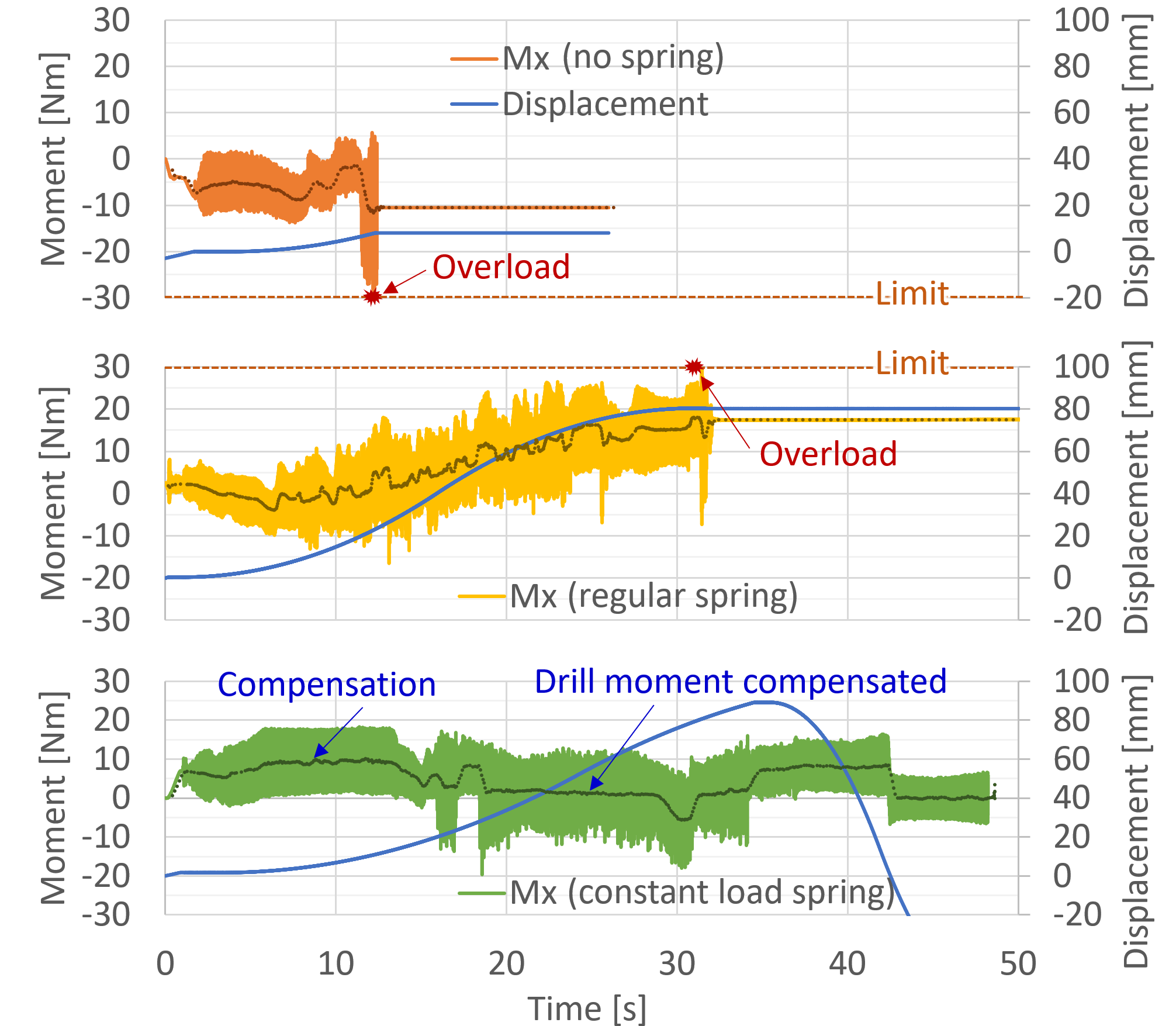}
  \caption{Hole drilling results. Top: No moment compensation. Middle: Compensation with regular spring. Bottom: Compensation with constant load spring.}
  \label{fig:res_drill}
  \end{center}
\end{figure}

The evaluation results are shown in Fig. \ref{fig:res_drill}. Because the drill of all tested tools were placed offset from the robot flange, the moment overload caused by the perpendicularity error shown in Fig. \ref{subfig:drillconcept1} does not occur. As shown in the figure, the hole could not be drilled when using the drilling tool without compensation or with the regular spring (drilling tool 1), but it could be successfully carried out by using the drilling tool with the constant load spring (drilling tool 2). This is because, without the support arm, the negative moment caused by the drill in the X axis (Mx) was not compensated, which caused the negative moment overload after advancing only 10 mm . For drilling tool 1, Mx gradually increased the deeper the hole was drilled due to the spring compression, which led to positive moment overload from moment overcompensation. In drilling tool 2, these problems did not occur because the Mx increased at the beginning due to the contact of the support arm to the wall, and then it did not increase as the hole was drilled. This behavior was expected because the constant load spring applies a constant force independent of the distance it is pulled . This result demonstrates that the proposed drilling tool design enables hole drilling with shorter tool length, appropriately compensates for the moment caused by the offset of the drill used, and reduces the reaction force applied to the robot so that smaller robots can be used to drill holes in concrete.

\subsection{Anchor hammering measurement}
To evaluate the anchor hammering tool, anchor hammering was attempted with the tool. An anchor was grasped with the rubber gripper of the tool and then inserted into a hole opened in the concrete block until the moment in the X axis reached an \textit{insertion end threshold} of 90 Nm. In this condition, the anchor was inserted about 7 mm, but the anchor wedge caused the anchor to be stuck because its diameter was slightly larger than the hole. After that, the hammer tool was turned on, and the robot was moved forward with position control to attempt to hammer the anchor until it reached the bottom of the hole (about 80 mm deep). The robot was programmed to stop if an absolute moment over  the \textit{hammering end threshold} of 27 Nm was detected. The anchor hammering was considered successful if a hole depth over 70 mm and an absolute moment over the threshold was reached, which indicates that the anchor touched the bottom of the hole. If moments over the limit were detected, the safety function of the robot controller would automatically stop the robot.

\begin{figure}[t!]
  \begin{center}
  \includegraphics[width=\columnwidth]{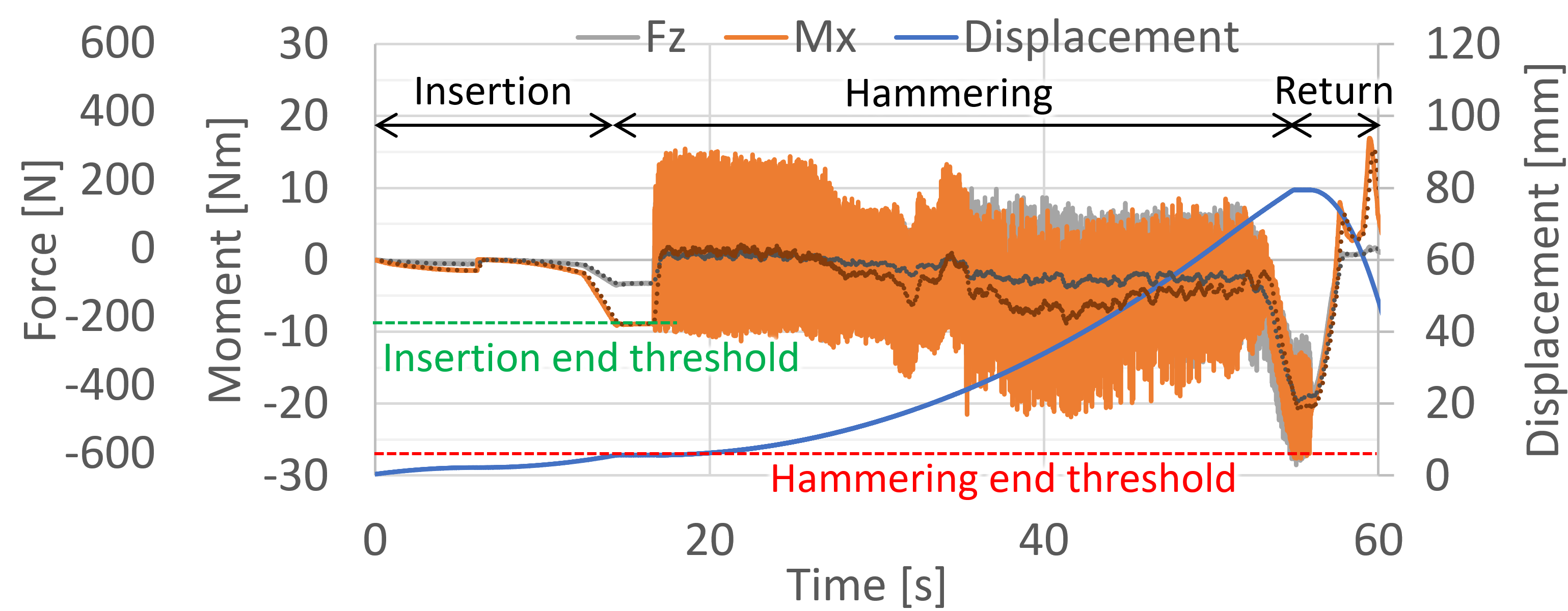}
  \caption{Anchor insertion and hammering results (Step 7 and 8 of fixing procedure shown in Fig. \ref{fig:procedure}, respectively).} %\textit{Start} is the threshold to detect the anchor was inserted until the maximum allowed depth without hammering. \textit{Finish} is the threshold to detect the anchor touched the bottom of the hole.}
  \label{fig:res_hammer}
  \end{center}
\end{figure}

As shown in the evaluation results in Fig. \ref{fig:res_hammer}, the anchor hammering was successful as the moment did not surpass the \textpm30 Nm limit, robot displacement reached about 75 mm, and the moment decreased suddenly when it reached the moment threshold. This suggests that anchor hammering can be adequately conducted with the proposed anchor hammering tool even with smaller robots.

\begin{figure}[t!]
  \begin{center}
  \includegraphics[width=\columnwidth]{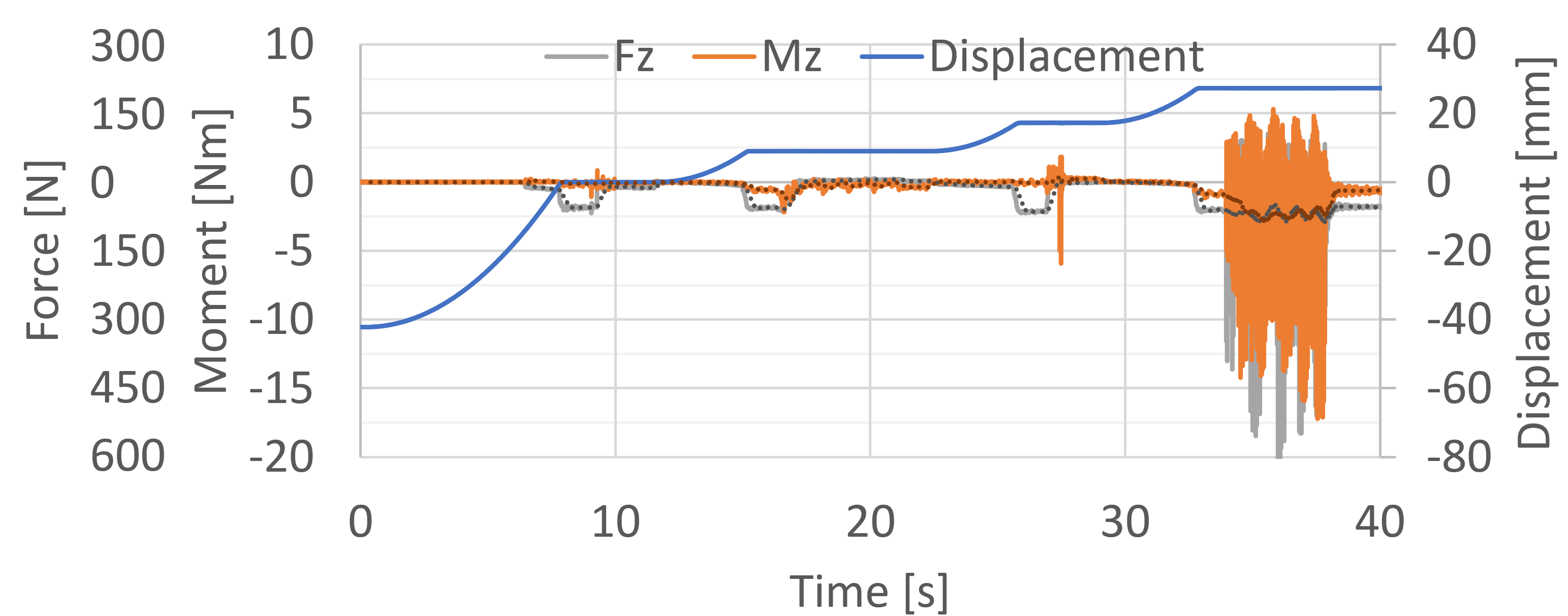}
  \caption{Nut tightening results.}
  \label{fig:res_nut}
  \end{center}
\end{figure} 

\begin{figure*}[t!]
  \begin{center}
  \includegraphics[width=\linewidth]{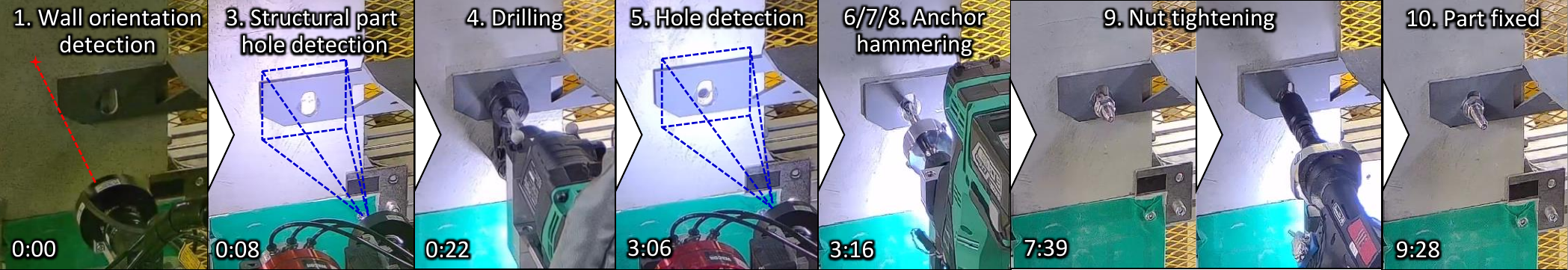}
  \caption{Screenshots of the support object fixing procedure.}
  \label{fig:res_cont_task}
  \end{center}
\end{figure*}

\subsection{Nut tightening}
To evaluate the nut tightening tool, the anchor was tightened in the following steps: (1) approaching the anchor bolt until a force measurement in the Z axis reached the threshold of 50 N; (2) rotating the socket rapidly alternating between clockwise and counter-clockwise to fit the nut into the socket (detected when force suddenly decreases and the socket spring is extended); (3) advancing the tool until the threshold is reached again, (4) rotating the nut until the nut reaches the surface of the structural part, (5) advancing the tool again until the threshold is reached, and (6) tightening the nut with a torque of 50 Nm.

The results of the nut tightening with the above procedure is shown in Fig. \ref{fig:res_nut}. As shown, the nut tightening was successfully carried out as the moment limit of \textpm30 Nm was not reached. This indicates that the tool design and the pulse mechanism of nut runner selected sufficiently reduces the reaction force in the robot arm so that the nut can be tightened with high torque without moment overload.

\subsection{Structural part fixation}
The proposed structural part fixation procedure was evaluated in the implemented experimental setup. Even though pick and place is also included in the procedure, for this test it was assumed that robot 2 had already picked the structural part and placed it on the target position on the wall. Screenshots of the task being executed are shown in Fig. \ref{fig:res_cont_task}. Details of the task execution are shown in the supplementary video. As shown, one side of the structural part could be adequately fixed with the proposed procedure in 9 min 28 s. The tasks that took the longest were the drilling (2 min 44 s), anchor hammering (4 min 28 s), and nut fastening (1 min 49 s) as they included tool attachment and detachment. Anchor hammering was specially long because the anchor insertion did not succeeded in the first insertion attempt, and a hole search based on a spiral trajectory had to be conducted to find the hole, which lasted about 60 s. To reduce the task execution time, the robot speed during tool change could be increased, and a deep learning-based hole search algorithm such as the shown in \cite{andre_icra2021} could be used.

\section{CONCLUSION}
In this study, we proposed a robot system for automating the complete fixation of structural parts to concrete surfaces. The proposed system is able to fix structural parts due to its dual-arm structure. In addition, we developed custom tools that enable the usage of smaller robots with lower payloads in narrow environments. We also proposed a method of transporting the system in parts to the work environment and a structural part fixation procedure. Experimental results show that the proposed robot system, tool designs, and fixation procedure enable the robot to successfully fix a structural part automatically. Although the proposed system is still a concept and has not been extensively tested, the results show its potential for application in the construction field. The fixation procedure of this study is still fixed, but a study has been conducted to change it dynamically to recover from failures \cite{hamasaki_sii2023}.

\addtolength{\textheight}{-10cm}   % This command serves to balance the column lengths


% \section*{ACKNOWLEDGMENT}

\bibliographystyle{IEEEtran}
\bibliography{myBib}

\end{document}